\documentclass[conference]{IEEEtran}
\IEEEoverridecommandlockouts

\usepackage{cite}
\usepackage{amsmath,amssymb,amsfonts}
\usepackage{algorithmic}
\usepackage{graphicx}
\usepackage{textcomp}
\usepackage{xcolor}
\usepackage{booktabs}
\usepackage{multirow}
\usepackage{hyperref}
\usepackage{url}
\usepackage{orcidlink} 

\def\BibTeX{{\rm B\kern-.05em{\sc i\kern-.025em b}\kern-.08em
    T\kern-.1667em\lower.7ex\hbox{E}\kern-.125emX}}

\usepackage{lipsum}
\usepackage{fancyhdr}
\fancypagestyle{sec}{\lhead{\tmpx}}

\fancypagestyle{arXiv}
{
   \fancyhf{}
   \chead{\textcolor{gray}{This paper is accepted for presentation at the  2026 IEEE International Conference on Metrology for eXtended Reality, Artificial Intelligence and Neural Engineering - IEEE MetroXRAINE 2026, Chemnitz, Germany}}
   \cfoot{ \fontsize{=9pt}{10pt}\selectfont  \textcolor{gray}{© 2026 IEEE. Personal use of this material is permitted. Permission from IEEE must be obtained for all other uses, in any current or future media, including reprinting/republishing this material for advertising or promotional purposes, creating new collective works, for resale or redistribution to servers or lists, or reuse of any copyrighted component of this work in other works}\\\fontsize{=10pt}{12pt}\selectfont\thepage}
   
}

\begin{document}

\title{Which Metric Reflects the Spelling Rate Accuracy in Event-Related Potential--Based Brain--Computer Interfaces?}

\author{\IEEEauthorblockN{Okba Bekhelifi \orcidlink{0000-0001-8629-2240}}
\IEEEauthorblockA{\textit{Intelligent Systems Research Laboratory (LARESI)} \\
\textit{USTOMB}\\
Oran, Algeria \\
okba.bekhelifi@univ-usto.dz}
\and
\IEEEauthorblockN{Naoual El Djouher Mebtouche \orcidlink{0000-0002-3855-5437}}
\IEEEauthorblockA{\textit{LRIA, Faculty of Computer Science} \\
\textit{USTHB}\\
Algiers, Algeria \\
nmebtouche@usthb.dz}
}

\maketitle

\begin{abstract}
For predictive models, the often-reported performance metrics are the loss and accuracy. In synchronous Brain-Computer Interface (BCI) systems, these metrics are informative for most BCI paradigms; however, for Event-Related Potential (ERP) applications the spelling rate, which measures the number of characters correctly selected is more important as it influences the estimation of information transfer rate (ITR) and any related metric measuring spelling performance. Moreover, ERP-based BCIs hold imbalanced data class distributions, which require reporting metrics that can handle the imbalance, such as the area under the receiver operating characteristic curve (ROC AUC). In this work, we study the correlation of the spelling rate with 13 metrics to identify  which among them best reflect user spelling performance and how they are affected by trial repetition. The Results of two datasets (a private LARESI ERP dataset and the public OpenBMI ERP dataset) favor the Brier score, Matthews Correlation Coefficient (MCC), and the metrics that account for class imbalance in binary classification: ROC AUC, area under the Precision-Recall curve (PR AUC), Average Precision (AP), and partial AUC (pAUC). These findings encourage researchers and practitioners to report those metrics in ERP-based BCI experiments.
\end{abstract}


\section{Introduction}
\thispagestyle{arXiv}
With rapid progress in the Brain-Computer Interface (BCI) field, invasive systems that implant electrodes into brain tissue can restore communications for amyotrophic lateral sclerosis (ALS) patients~\cite{card2024, kunz2025}. Their surgical-free alternatives use Electroencephalography (EEG) to capture and translate brain signals into digital commands. The standard EEG-based BCI paradigms either rely on voluntary modulation of sensorimotor cortical brainwaves by Motor Imagery (MI)~\cite{cho2018} or on the cortical response evoked by attending to a stimulus as in Event-Related Potentials (ERPs)~\cite{kalra2023} and Steady State Visual Evoked Potentials (SSVEPs)~\cite{volosyak2020}.

As a nascent field, despite explosive growth in research and efforts to set rigorous methods for design and evaluation, BCI still needs standards for performance and usage reporting. Often, researchers report their results differently, as the experimental design space is large for unrelated paradigms. Additionally, EEG is inherently non-stationary, and acquisition systems, subjects, and sessions carry significant variability~\cite{melnik2017}. Nevertheless, many guidelines and tutorials have been introduced to address this situation. A step-by-step guide for designing a rigorous BCI experiment was presented in~\cite{jeunet2018}. One of the first studies to review BCI performance reporting~\cite{billinger2012} listed confusion matrix, accuracy, Cohen's Kappa, Sensitivity, Specificity, receiver operator characteristic (ROC) curve, F-Measure, Information Transfer Rate (ITR), Written Symbol Rate (WSR) and Practical Bit Rate (PBR) as performance measures. A similar study~\cite{thompson2013} divided the metrics into two levels: level~1 for classifier output measures and level~2 for translating logical control into semantic control; the latter added system efficiency and BCI-Utility metric. To concretize community consensus, checklists for methods reporting were introduced in~\cite{thompson2014} for different types of BCI research, adding user-centric metrics such as cognitive workload and usability.

ERP-based BCIs use the oddball paradigm to present a set of stimuli to the user sequentially while recording the EEG response to each stimulus simultaneously. The target stimulus appearance is rare against other non-target events. When the subject attends to the target stimulus, particular negative and positive peaks appear at distinct intervals after stimulus onset; these peaks are called ERPs. To construct a communication system with ERPs, alphanumeric characters are typically arranged in a $6\times6$ grid forming a speller. The stimulation goes on by flashing entire rows and columns in a random order with multiple repetitions. Stimulus presentation is not limited to flashing characters; face presentation has been adopted due to its superiority in eliciting more pronounced and robust ERPs~\cite{kaufmann2013}. Row/column presentation has also been improved to reduce errors caused by flashing  of neighboring character~\cite{yeom2014}.

The most commonly applied metric to assess online BCI performance is the ITR, which estimates the amount of information per time unit generated by the BCI. The seminal study~\cite{yuan2013} analyzed the pitfalls in reporting ITR when prerequisites for correct estimation are not met and proposed recommendations to ensure accurate calculation. The ITR depends on three variables: ($N$)~the number of available choices, ($T$)~the duration of time to output a single command, and ($P$)~the accuracy of the online test.

The original formulation of ITR ignores the realistic operation of a speller where the system produces an erroneous character. Improvements accounting for correction were introduced as Word Symbol Rate (WSR)~\cite{furdea2009} and Practical Bit Rate (PBR)~\cite{townsend2010}. Other metrics address the limitations of ITR: Efficiency~\cite{quitadamo2012} considers classification accuracy, cost of errors, and correction strategies; Average Time Consumption Per Character (ATCPC)~\cite{wang2024} measures the mean time spent to successfully type a character. Further studies examined spelling rate estimation under the effect of latency jitter~\cite{thompson2013b} and for projecting long-term accuracy~\cite{colwell2014}.

In ERP-based BCI, $P$ is referred to as the \emph{spelling rate}; it calculates the percentage of correctly spelled characters. Character selection involves a two-stage operation: all events are classified with a score of being the target, then the event with the maximum score is taken as the character. Often, event trials are repeated and scores averaged to improve accuracy. Consequently, classifier performance metrics and spelling rate are not directly related, which raises the question: \emph{which metric best corresponds to the spelling rate in these settings?} This study evaluates the correlation between spelling rate ($P$) and 13 binary classification metrics to identify which indicators best reflect functional performance. Furthermore, we investigate parameters that affect this relationship, showing that metrics accounting for the class imbalance are highly correlated with spelling rate and that repetition reduces correlation power.

\section{Materials and Methods}

\subsection{Metrics}

Given $n$ EEG epochs $X_i \in \mathbb{R}^{C \times S}$ with true labels $y_i \in \{0, 1\}$, $i = 1\ldots n$, where $C$ is the number of EEG channels and $S$ is the number of time samples. The output of a classifier for an input $X_i$ is $\hat{y_i}$ . In a binary classification setting as ERP classification of target vs non-target classes, the possible outcomes of the classifier are split into four categories: True positives (TP) epochs where the true and predicted labels are 1, True Negatives (TN) epochs where the true and predicted labels are 0, False Positives (FP) epochs where the true label is 0 and the predicted label is 1, and False Negatives (FN) epochs where the true label is 1 and the predicted label is 0. ~\cite{varoquaux2023}.

With these fractions, different metrics are defined to measure the classifier’s performance in determining the true target and non-target epochs to increase \emph{spelling rate} $P$ , defined as:
\begin{equation}
P = \frac{1}{N_c}\sum_{j=1}^{N_c} \mathbf{1}\!\left(\widehat{\mathrm{Char}}_j = \mathrm{Char}_j\right) \label{eq:spellingrate}
\end{equation}
\begin{equation}
\widehat{\mathrm{Char}}_j = \arg\max_{k} O_{jk},\quad k = 1\ldots n_j \label{eq:argmax}
\end{equation}
where $N_c$ is the number of characters to spell in a session, $n_j$ is the number of epochs per character spelling trial, $\mathrm{Char}_j$ is the true character, $\widehat{\mathrm{Char}}_j$ is the detected character, $\mathbf{1}(\cdot)$ is the indicator function, and $O_{jk}$ is the classifier output probability for the $k$-th event epoch in the $j$-th set, computed as the mean of scores for each stimulus.

The investigated metrics are divided into two categories: (1)~model performance metrics for imbalanced binary classification, and (2)~metrics that assess the model’s probability outputs alignment with the true labels. In the first category, we selected 10 metrics that are defined in the following subsection

\subsubsection{Model Performance Metrics}

\textbf{Precision} (Positive Predictive Value, (PPV)) ~\cite{varoquaux2023}measures the proportion of predicted positive epochs that are truly positive. A high Precision indicates that when the model predicts positive, it is usually correct: 
\begin{equation}
\text{Precision} = \frac{\mathrm{TP}}{\mathrm{T²P}+\mathrm{FP}}
\end{equation}

\textbf{Recall} (Sensitivity, True Positive Rate (TPR)) ~\cite{varoquaux2023} measures the proportion of actual positive epochs correctly identified by the model. A high Recall means the model identifies most positive cases:
\begin{equation}
\text{Recall} = \frac{\mathrm{TP}}{\mathrm{TP}+\mathrm{FN}}
\end{equation}

\textbf{Balanced Accuracy (BA)}~\cite{brodersen2010} is a metric designed for imbalanced datasets. It is computed as the arithmetic mean of sensitivity and specificity (recall for the negative class). Unlike standard accuracy, balanced accuracy gives equal weight to both classes regardless of their relative frequencies. A score of 0.5 corresponds to random chance; 1.0 indicates perfect classification:
\begin{equation}
\text{BA} = \frac{1}{2}\!\left(\frac{\mathrm{TP}}{\mathrm{TP}+\mathrm{FN}} + \frac{\mathrm{TN}}{\mathrm{TN}+\mathrm{FP}}\right)
\end{equation}

\textbf{F1-Score}~\cite{chicco2020} is the harmonic mean of Precision and Recall. It provides a single summary metric that balances both false positives (captured by Precision) and false negatives (captured by Recall). The F1 score ranges from 0 (worst) to 1 (best):
\begin{equation}
\text{F1} = \frac{2\,\text{Precision}\times\text{Recall}}{\text{Precision}+\text{Recall}} = \frac{2\mathrm{TP}}{2\mathrm{TP}+\mathrm{FP}+\mathrm{FN}}
\end{equation}

\textbf{Matthews Correlation Coefficient (MCC)}~\cite{chicco2020} considers all four outcomes of binary classification (TP, TN, FP, FN). MCC ranges from $-1$ (perfect inverse prediction) to $+1$ (perfect prediction), with 0 indicating random prediction:
\begin{equation}
\text{MCC} = \frac{\mathrm{TP}\!\cdot\!\mathrm{TN}-\mathrm{FP}\!\cdot\!\mathrm{FN}}{\sqrt{(\mathrm{TP}+\mathrm{FP})(\mathrm{TP}+\mathrm{FN})(\mathrm{TN}+\mathrm{FP})(\mathrm{TN}+\mathrm{FN})}}
\end{equation}

\textbf{Area Under the Receiver Operating Characteristic Curve}~\cite{fawcett2006} (ROCAUC) is formed by plotting the TPR (sensitivity) against the False Positive Rate (FPR) (the proportion of actual negative cases that are incorrectly identified as positive) at all classification thresholds. It measures a model's ability to discriminate between positive and negative classes across all thresholds.
An AUC of 0.5 indicates no discriminative ability (random guessing), while an AUC of 1.0 indicates perfect discrimination. Probabilistically, it equals the probability that a randomly chosen positive instance is ranked higher than a randomly chosen negative instance
s:
\begin{equation}
\text{ROCAUC} = \int_0^1 \mathrm{TPR}\,d(\mathrm{FPR})
\end{equation}

\textbf{Area Under the Precision-Recall Curve}~\cite{saito2015} (PRAUC)  is the area under the curve formed by plotting Precision against Recall at varying classification thresholds, where P(R)  denotes precision as a function of recall. A high PRAUC indicates that the model achieves both high precision and high recall. The baseline for a random classifier equals the proportion of positives in the dataset:
\begin{equation}
\text{PRAUC} = \int_0^1 P(R)\,dR
\end{equation}

A similar metric \textbf{Average Precision (AP)} ~\cite{zhu2004} was also used since it differs from PRAUC in the calculation method. AP is a weighted mean, and the PR AUC is interpolated using the trapezoidal rule:
\begin{equation}
\text{AP} = \sum_n P_n(R_n - R_{n-1})
\end{equation}
where $P_n$ and $R_n$ are precision and recall at the $n$-th threshold.

\textbf{Area Under Precision-Recall-Gain curves}~\cite{flach2015}(AUPRG) addresses a fundamental issue in standard PR analysis: the area under a traditional PR curve takes the arithmetic mean of precision values, while the F1 score uses the harmonic mean, making PRAUC a poor proxy for expected F1 performance. AUPRG is defined by re-expressing Precision and Recall as gains relative to a random classifier with class prior $\pi = (\mathrm{TP}+\mathrm{FN})/n$:
\begin{align}
\text{PrecisionGain} &= \frac{\text{Precision}-\pi}{(1-\pi)\cdot\text{Precision}} \\[4pt]
\text{RecallGain} &= \frac{\text{Recall}-\pi}{(1-\pi)\cdot\text{Recall}} \\[4pt]
\text{AUPRG} &= \int_0^1 \text{PrecisionGain}\,d(\text{RecallGain})
\end{align}

\textbf{Partial AUC (pAUC)}~\cite{dodd2003} is the area under the ROC curve restricted to a specific range $[a,b]$ of FPR values. It is used when only a relevant portion of the ROC curve is of interest. The pAUC can be standardized by dividing by (b-a) to yield values between 0 and 1 :
\begin{equation}
\text{pAUC}(a,b) = \int_a^b \mathrm{TPR}(t)\,dt
\end{equation}
We calculated pAUC for intervals $[0, b_i]$ with $b_i \in \{0.1, 0.2, 0.3, 0.4, 0.5\}$.
The remaining 3 metrics are defined below.
\subsubsection{Model Probability Output Metrics}

\textbf{Binary Cross-Entropy (BCE)}~\cite{baum1987} (log loss) measures the difference between the distributions of both predicted probabilities and true binary labels, penalizing confident but wrong predictions:
\begin{equation}
\text{BCE} = -\frac{1}{n}\sum_{i=1}^n \bigl[y_i\log\hat{y}_i + (1-y_i)\log(1-\hat{y}_i)\bigr]
\end{equation}

\textbf{Brier Score}~\cite{brier1950} measures the mean squared difference between predicted probabilities and actual binary outcomes. It ranges from 0 (perfect predictions) to 1 (worst). The Brier score evaluates both calibration and discrimination simultaneously and can be decomposed into reliability, resolution, and uncertainty components:
\begin{equation}
\text{Brier} = \frac{1}{n}\sum_{i=1}^n (y_i - \hat{y}_i)^2
\end{equation}

\textbf{Expected Calibration Error (ECE)}  ~\cite{guo2017} measures how well a model's predicted probabilities align with true empirical outcome frequencies. Predictions are grouped into $M$ bins; $|B_m |$ is the number of samples in bin $m$, $acc(B_m)$ is the accuracy, and $conf(B_m)$ is the mean predicted probability in that bin. An ECE value of 0 indicates perfect calibration:
\begin{equation}
\text{ECE} = \sum_{m=1}^M \frac{|B_m|}{n}\bigl|\text{acc}(B_m)-\text{conf}(B_m)\bigr| \\
\end{equation}
\textbf{Maximum Calibration Error (MCE)}  ~\cite{guo2017}  identifies the most severely miscalibrated region, Unlike ECE, which averages miscalibration across bins. It is particularly relevant in high-stakes applications where worst-case reliability is critical
\begin{equation}
\text{MCE} = \max_{m\in\{1,\ldots,M\}}\bigl|\text{acc}(B_m)-\text{conf}(B_m)\bigr|
\end{equation}

\subsection{Datasets}

\textbf{Dataset~I (LARESI ERP):} The ERP dataset from a Hybrid BCI study~\cite{bekhelifi2024}. Six subjects participated with 15 EEG g.Sahara dry electrodes sampled at 512~Hz. Nine directional icons were flashed using a single character flashing scheme with stimulation duration of 70~ms and inter-stimulus interval (ISI) of 30~ms. Two runs were conducted per subject: in the calibration run, the flashing repetition was set to 10; in the online run, the repetition was 1 (single-trial). Each session comprised 36 randomly generated characters to spell. The class imbalance ratio was 1:8.

\textbf{Dataset~II (OpenBMI ERP):} From the large OpenBMI three-paradigm dataset~\cite{lee2019}. We selected the ERP subset with 54 participants across two sessions. Each session included calibration (33 characters) and online test (36 characters) runs. The speller was a conventional $6\times6$ grid flashed with familiar faces following the random-set presentation scheme~\cite{yeom2014}, with stimulation duration of 80~ms, ISI of 135~ms, and 5 flashing (sequence) repetitions. 62 Ag/AgCl electrodes collected EEG at 1000~Hz, downsampled to 250~Hz. We selected 15 channels matching the LARESI ERP dataset, replacing PO7/PO8 with P7/P8. The class imbalace ratio was 1:5.

\subsection{Evaluation}

EEG data were filtered with a 2nd-order IIR Butterworth bandpass filter in $[1,\,12]$~Hz. Epochs were segmented in the interval $[0,\,600]$~ms following stimulus markers and baseline-corrected using 300~ms prior to stimulus onset. Two classification methods were evaluated: (1)~a classical spatial filtering pipeline comprising xDAWN~\cite{rivet2009} with 3 components followed by Logistic Regression; and (2)~the convolutional neural network EEG-TCNet~\cite{ingolfsson2020}, comprising three convolutional blocks, two Temporal Convolution Network (TCN) blocks, and a linear layer. Both methods were selected for their proven high performance on ERP datasets in single-subject evaluation, and for minimizing BCE as a loss function. EEG-TCNet was trained for 500 epochs using the AdamW optimizer (learning rate $10^{-3}$, weight decay $10^{-4}$, batch size 64). Training was run for five different seeds and results are reported as the mean over these runs.

Statistical significance analysis between Logistic Regression and EEG-TCNet followed this procedure: Shapiro--Wilk normality test, followed by Wilcoxon signed-rank test if normality failed, and Cohen's~$d$ for effect size. The correlation between spelling rate ($P$) and 18 metrics was assessed by Pearson's correlation coefficient~$r$ (linear relationship) and Kendall's~$\tau$ (ordinal association). The significance level was adjusted using Bonferroni correction to $\alpha = 0.0027$ for model comparison and to $\alpha = 0.0028$  for correlation analyses.

\begin{table}[!t]
\centering
\caption{Correlation analysis results for the LARESI ERP dataset. Significance codes: $^{**}p<0.001$, $^{*}p<0.0028$, $^{\dagger}$marginal $p=0.0028$. LogReg refers to Logistic Regression}
\label{tab:laresi_corr}
\renewcommand{\arraystretch}{1.2}
\setlength{\tabcolsep}{4pt}
\begin{tabular}{lcccc}
\toprule
 & \multicolumn{2}{c}{\textbf{Pearson's $r$}} & \multicolumn{2}{c}{\textbf{Kendall's $\tau$}} \\
\cmidrule(lr){2-3}\cmidrule(lr){4-5}
\textbf{Metric} & LogReg & EEG-TCNet & LogReg & EEG-TCNet \\
\midrule
BCE          & $-0.9209$         & $-0.8535$         & $-0.8667$        & $-0.7333$       \\
Precision    & $0.9272$          & $0.8950$          & $0.8667$         & $0.8667$        \\
Recall       & $0.9549$          & $0.9430$          & $0.9309$         & $0.6901$        \\
\textbf{BA}  & $0.9731^{*}$      & $0.9638^{*}$      & $0.8667$         & $0.8667$        \\
ROCAUC          & $0.8212$          & $0.7479$          & $0.8667$         & $0.7333$        \\
\textbf{PRAUC} & $0.9839^{**}$ & $0.9557$          & $0.8667$         & $0.8667$        \\
AUPRG        & $0.9270$          & $0.8413$          & $0.8667$         & $0.7333$        \\
\textbf{AP}  & $0.9831^{**}$     & $0.9555$          & $0.8667$         & $0.8667$        \\
\textbf{pAUC ($b$=0.1)} & $0.9858^{**}$ & $0.9820^{**}$ & $1.0^{\dagger}$ & $0.8667$ \\
\textbf{pAUC ($b$=0.2)} & $0.9704^{*}$  & $0.9525$      & $0.8667$         & $0.8667$        \\
pAUC ($b$=0.3) & $0.9408$        & $0.9167$          & $0.8667$         & $0.8667$        \\
pAUC ($b$=0.4) & $0.9105$        & $0.8733$          & $0.8667$         & $0.7333$        \\
pAUC ($b$=0.5) & $0.8835$        & $0.8339$          & $0.8667$         & $0.7333$        \\
\textbf{F1}  & $0.9573^{*}$      & $0.9847^{**}$     & $0.8667$         & $1.0^{\dagger}$ \\
\textbf{Brier} & $-0.9489$       & $-0.9656^{*}$     & $-0.8667$        & $-1.0^{\dagger}$\\
\textbf{MCC} & $0.9633^{*}$      & $0.9829^{**}$     & $0.8667$         & $1.0^{\dagger}$ \\
\textbf{ECE} & $0.9817^{**}$     & $0.9722^{*}$      & $0.8667$         & $1.0^{\dagger}$ \\
MCE          & $0.8064$          & $0.8530$          & $0.7333$         & $0.6000$        \\
\bottomrule
\end{tabular}
\end{table}

\begin{figure*}[!t] 
  \centering
  
  \begin{minipage}{\textwidth}
    \centering
    \includegraphics[width=0.9\textwidth]{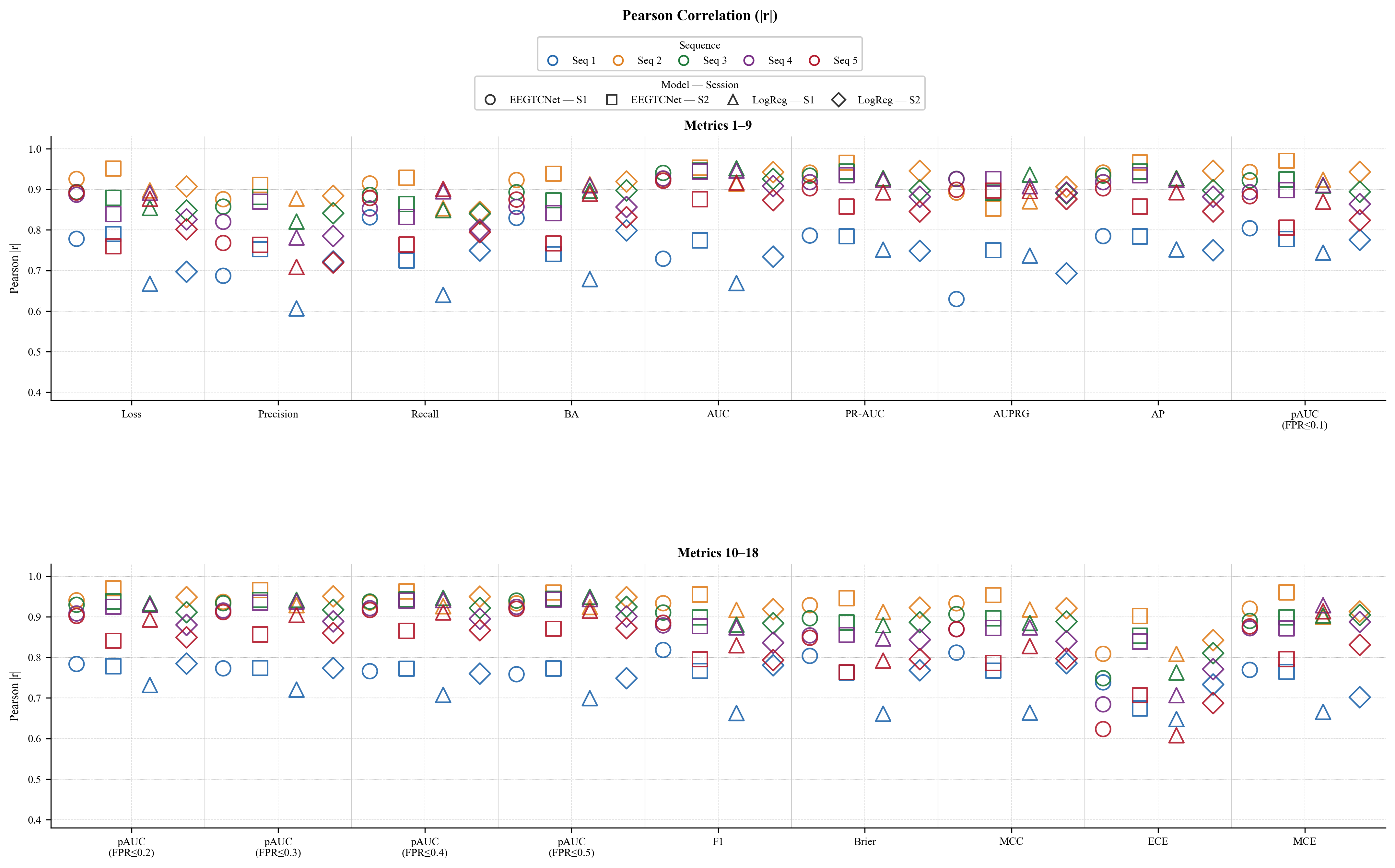}
    \caption{Pearson’s $r$ correlation coefficient values for OpenBMI ERP per repetition sequence. Absolute values are plotted. Circles and square represent EEGTCNet session 1 and session 2 respectively.  LogReg refers to Logistic Regression and is represented by triangle and diamond for session 1 and session 2 respectively}
    \label{fig:openbmi_pearson}
  \end{minipage}


  \begin{minipage}{\textwidth}
    \centering
    \includegraphics[width=0.9\textwidth]{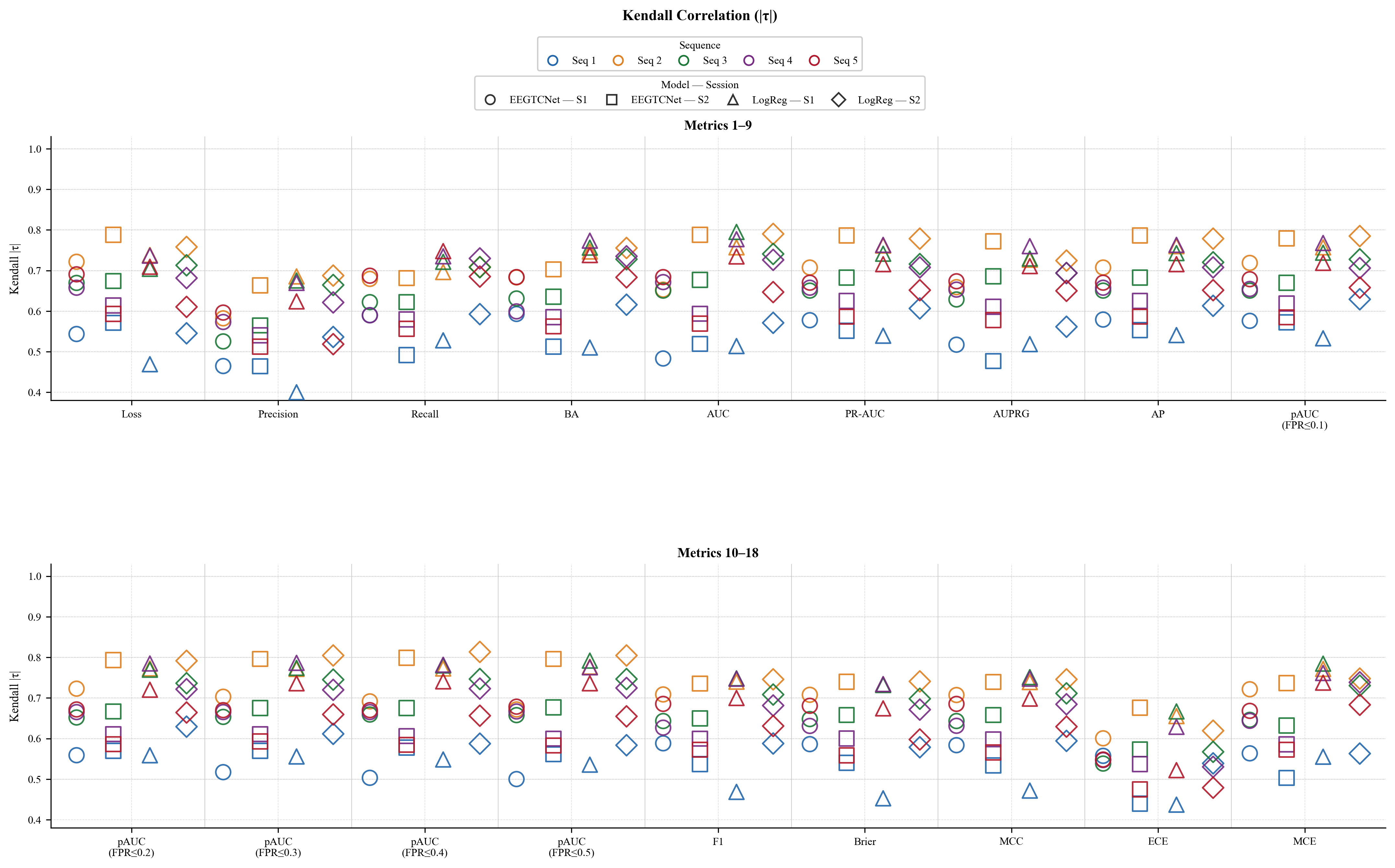}
    \caption{Kendall’s $\tau$ correlation coefficient values for OpenBMI ERP per repetition sequence. Absolute values are plotted. Absolute values are plotted, Circles and square represent EEGTCNet session 1 and session 2 respectively.  LogReg refers to Logistic Regression and is represented by triangle and diamond for session 1 and session 2 respectively.}
    \label{fig:openbmi_kendall}
  \end{minipage}

  \label{fig:whole_page_figures}
\end{figure*}

\section{Results}
The classifier comparison did not reveal any significant difference for the LARESI ERP dataset. For OpenBMI ERP session~1 at one repetition sequence, the metrics Brier, BCE loss, ECE, and Precision showed large effect, while AP, $P$, F1, MCC, pAUC ($b = 0.1$, $b = 0.2$), and PRAUC showed medium effect. Metrics ROCAUC, AUPRG, and pAUC ($b = 0.3, 0.4, 0.5$) revealed small effect, and BA a negligible effect; Recall was non-significant. The same patterns held for two, three, four, and five repetitions, with differences in effect size. For OpenBMI session~2, the same patterns were observed across metrics and sequence repetitions. All Wilcoxon signed-rank results for OpenBMI were significant with $p < 0.001$.

The correlation results for LARESI ERP are presented in Table~\ref{tab:laresi_corr}. For Pearson's~$r$, the metrics BA, pAUC ($b = 0.1$), F1, MCC, and ECE showed high correlation values ($r > 0.957$) with significance in both models ($p < 0.0028$ or $p < 0.001$). PRAUC and pAUC ($b = 0.2$) showed significance only for Logistic Regression, while Brier was significant only in EEG-TCNet. For Kendall's~$\tau$, marginal significance ($p = 0.0028$) was observed in pAUC ($b = 0.1$) for Logistic Regression, and for EEG-TCNet the marginally significant metrics were F1, Brier, MCC, and ECE.

The OpenBMI Pearson's~$r$ results are depicted in Fig.~\ref{fig:openbmi_pearson}. All metrics had moderate to strong correlation at one repetition ($0.6 \leq r \leq 0.8$) in both sessions. The consistent metrics with very strong correlation for the remaining repetitions in session~1 were ROCAUC, PRAUC, AUPRG, AP, and pAUC (all five $b$ values). In session~2, the same metrics maintained this pattern only for two and three repetitions. Kendall's~$\tau$ results for OpenBMI are shown in Fig.~\ref{fig:openbmi_kendall}: one repetition had weak to moderate correlation ($0.5 \leq \tau \leq 0.6$) in most metrics in both sessions. The same metrics as in Pearson's~$r$ exhibited higher correlation for two, three, and four repetitions, with a more stable strong Kendall's~$\tau$ ($\tau \approx 0.8$) for two repetitions, especially in session~2. All $p$-values for the correlation analysis on the OpenBMI dataset were significant ($p < 0.001$).

\section{Discussion}

In the model comparison, probability output metrics along with Precision carried the highest discriminative power between models, followed by metrics that weight true positives and false positives more heavily than false negatives. This indicates that both models differ significantly in handling false positives while being equivalent in handling true positives and false negatives. In addition, it confirms that the neural network's predicted probabilities align better with ground truth.

The linear relationship of spelling rate with metrics (Pearson's~$r$) was stable except for Precision and ECE, which showed the weakest correlation. Metrics revealed identical or near-identical patterns forming sets: $\{$F1, MCC, Brier$\}$, $\{$PRAUC, AP$\}$, $\{$Recall, BA$\}$; BCE loss held the strongest correlation. Consistent sets were more pronounced for EEG-TCNet, suggesting higher robustness in convolutional networks. Although AUPRG assumes a relationship to F1, the results do not support this assumption. The ordinal association (Kendall's~$\tau$) mirrored the same phenomenon with identical metrics for EEG-TCNet in the same set $\{$F1, MCC, Brier$\}$, $\{$Recall, BA$\}$, and a set for threshold-based metrics $\{$ROCAUC, PRAUC, AUPRG, AP, pAUC$\}$. For both association types, the dominant factor was the repetition sequence: two repetitions yielded the strongest concordance.

This effect arises because spelling rate estimation takes a mean of probabilities for the same repeated stimulus, transforming binary classification into a multiple-choice decision. These non-linear operations skew scores towards target events with more repetitions, changing the nature of the relationship between spelling rate and per-instance metrics. A single trial can be sufficient for a speller with few characters and simple single-stimulus flashing, while two repetitions preserve the linear and monotonic associations for a more general speller with Row/Column or random set stimulation.

Surprisingly, the calibration metric ECE showed the weakest association. Although MCE ranked highly, Brier score stood out with high correlations, indicating that calibration is better measured by the latter. The identical profile of F1 and MCC points toward choosing only one; in line with~\cite{chicco2020}, which demonstrates the advantages of MCC over F1, we favor adoption of MCC. The area-under-curve metrics shared the same pattern and consistently exhibited the strongest link, except AUPRG. Consequently, pAUC ($b=0.1$), AP, and PRAUC are the recommended metrics to add alongside ROCAUC. Since AP and PR AUC are different computational methods for the same underlying metric, we favor AP for its robustness. Finally, Precision, Recall, and BA ranked lowest, undermining their utility as performance reporting measures.

\section{Conclusion}

In this study, we investigated the relationship between spelling rate and different classification metrics in ERP BCI experiments. We highlighted the metrics with the strongest relationship to spelling performance and showed that metrics explicitly considering the ratio between true positive rate and false positive rate are better indicators of spelling performance. As a result, we encourage the BCI community to adopt them in reporting ERP-based BCI experiment results especially PRAUC, AP, pAUC, in addition to the widely used ROCAUC. Future work will generalize the study to more performance measures beyond the spelling rate.

\bibliographystyle{IEEEtran}

\end{document}